\documentclass{article}

\usepackage[utf8]{inputenc}
\usepackage[T1]{fontenc}
\usepackage{fancyhdr}
\usepackage[normalem]{ulem}
\useunder{\uline}{\ul}{}
\usepackage{PRIMEarxiv}

\usepackage{booktabs}
\usepackage{amsfonts}
\usepackage{nicefrac}
\usepackage{multirow}
\usepackage{array}

\usepackage{graphicx}
\usepackage{caption}
\usepackage{multicol}
\usepackage{microtype}

\usepackage{url}
\usepackage{hyperref}

\title{\textnormal{Shakti-VLMs: Scalable Vision-Language Models for Enterprise AI}}
\author{%
    \textbf{Syed Abdul Gaffar Shakhadri}\\
    \small{Lead AI Developer}\\
    \small{SandLogic Technologies Pvt Ltd.}\\
    \small{\texttt{syed.abdul@sandlogic.com}}
    \and
    \textbf{Kruthika KR}\\
    \small{AI Researcher}\\
    \small{SandLogic Technologies Pvt Ltd}\\
    \small{\texttt{kruthika.kr@sandlogic.com}}
    \and
    \textbf{Kartik Basavaraj Angadi}\\
    \small{AI Developer}\\
    \small{SandLogic Technologies Pvt Ltd}\\
    \small{\texttt{kartik.angadi@sandlogic.com}}
}

\begin{document}
\maketitle

\begin{abstract}
We introduce Shakti VLM, a family of vision-language models in the capacity of 1B and 4B parameters designed to address data efficiency challenges in multimodal learning. While recent VLMs achieve strong performance through extensive training data, Shakti models leverage architectural innovations to attain competitive results with fewer tokens. Key advancements include QK-Normalization for attention stability, hybrid normalization techniques, and enhanced positional encoding. A three-stage training strategy further optimizes learning efficiency. Evaluations show that Shakti-Shakti-VLM-1B and Shakti-VLM-4B excel in document understanding, Visual Reasoning, OCR extraction, and general multimodal reasoning. Our results highlight that high performance can be achieved through model design and training strategy rather than sheer data volume, making Shakti an efficient solution for enterprise-scale multimodal tasks.
\end{abstract}


\section{Introduction}
Large Vision-Language Models (LVLMs) have emerged as a transformative force in artificial intelligence, seamlessly integrating vision and language understanding to enhance multimodal perception and reasoning. By capitalizing on recent advancements in Vision Transformers (ViTs)\cite{dosovitskiy2021imageworth16x16wordsvit} and Large Language Models (LLMs), these systems can interpret images, documents, and videos with remarkable textual comprehension. Yet, existing LVLMs often face challenges such as high computational costs, fine-grained visual perception limitations, extended context handling issues, and difficulties adapting to real-world data diversity.

Recent advancements in vision-language models (VLMs) have significantly improved AI-driven multimodal applications, demonstrating strong performance in image-text understanding, object recognition, and reasoning. Notable models like Qwen2VL\cite{wang2024qwen2vlenhancingvisionlanguagemodels}, Molmo\cite{deitke2024molmopixmoopenweights}, and SmolVLM\cite{smolvlm} have showcased impressive capabilities but rely on extensive training data to achieve high accuracy across diverse tasks. This dependency presents scalability challenges, particularly for enterprise applications that require efficient and adaptable solutions.

To address these limitations, we introduce Shakti-VLM, a family of lightweight yet high-performing vision-language models (Shakti-VLM-1B and Shakti-VLM-4B), optimized for enterprise-scale and edge deployments. Building on insights from large-scale open-source efforts such as Qwen2.5-VL and InternVL, Shakti models focus on efficiency rather than size alone, ensuring robust multimodal capabilities while maintaining computational feasibility.

Shakti models incorporate several architectural innovations that improve efficiency and generalization across multimodal tasks. Rather than merely increasing model size, Shakti-VLMs employ a hybrid normalization strategy, leveraging QK-Normalization\cite{henry2020querykeynormalizationtransformers} for stable attention mechanisms and enhanced positional encoding, ensuring faster convergence and robust performance even under limited data scenarios. These design choices make Shakti models highly effective for document parsing, OCR extraction, and chart interpretation, making them ideal for real-world enterprise pipelines.

Our training approach follows a three-stage methodology to maximize efficiency. First, we pretrain the decoder on extended-context text-only data, enabling strong language understanding before multimodal alignment. Next, we align vision and language representations using a frozen decoder, ensuring effective feature fusion without unnecessary computational overhead. Finally, we perform full model fine-tuning, incorporating instruction tuning, RLHF\cite{rlhf}, and DPO\cite{dpo}, optimizing the model for real-world multimodal applications.This structured approach maximizes data efficiency, achieving strong multimodal alignment with significantly lower training requirements.

Despite using significantly fewer training tokens than other VLMs (487 billion for Shakti-VLM-1B and 782 billion for Shakti-VLM-4B), both models demonstrate exceptional benchmark performance. Shakti-VLM-1B delivers balanced results across diverse multimodal tasks, particularly excelling in document and chart understanding, frequently outperforming larger models like SmolVLM-2.25B\cite{smolvlm}. Meanwhile, Shakti-VLM-4B surpasses state-of-the-art models, including Qwen2VL-7B\cite{wang2024qwen2vlenhancingvisionlanguagemodels} and MiniCPM-V-2.6-8B\cite{yao2024minicpm26}, on complex multimodal reasoning benchmarks. Furthermore, Shakti models exhibit strong generalization across visual question answering (VQA), mathematical reasoning, and long-form textual comprehension tasks, frequently matching or surpassing models with significantly more parameters.

By integrating scalable vision encoders, advanced attention mechanisms, and an optimized three-stage training process, Shakti-VLM models redefine efficiency in multimodal AI. Their strong performance across OCR, document understanding, and vision-language reasoning tasks establishes them as leading solutions in the evolving LVLM landscape, catering to real-world enterprise needs. Unlike conventional VLMs that demand vast computational resources, Shakti models are optimized for both enterprise-scale and edge deployments, ensuring a favorable balance between accuracy, memory footprint, and inference speed. We evaluate Shakti models on a broad range of multimodal benchmarks, including OCR tasks, document VQA, chart understanding, and general vision-language QA, demonstrating comparable or superior results against other models.

Key Features of Shakti-VLM Models

\begin{itemize}
    \item \textbf{Adoption of QK-Normalization} for improved stability and performance.
    \item \textbf{Hybrid Normalization Strategy}, combining Pre-LayerNorm in early layers with Post-LayerNorm using RMSNorm in later layers, ensuring an optimal balance between stability and efficiency.
    \item \textbf{Optimized three-stage training methodology}, allowing better performance across tasks with fewer training tokens.
    \item \textbf{Scalability across different deployment scenarios}, from enterprise-level document automation to edge computing applications requiring lightweight multimodal AI models.
\end{itemize}

\section{Related Work: }

Recent years have seen rapid progress in vision-language models (VLMs), driven by breakthroughs in architecture, scaling laws, and multimodal alignment techniques. These models are becoming central to tasks that require a seamless understanding of both visual and textual inputs, such as visual question answering, image captioning, document understanding, and OCR. This section highlights key developments in the field, with a focus on pioneering VLM families and their contributions to model efficiency, document processing, and training innovations. 

\subsection{Advancement in Vision Language Models}
Recent advancements in vision-language models (VLMs) have significantly expanded the capabilities of multimodal AI systems. Several notable model families have emerged, each with distinct architectural approaches and scaling strategies. 

The Qwen-VL\cite{bai2023qwenvlversatilevisionlanguagemodel} \cite{wang2024qwen2vlenhancingvisionlanguagemodels} series (Qwen-VL and Qwen2-VL) represents significant milestones in open-source VLM development. The original Qwen-VL\cite{bai2023qwenvlversatilevisionlanguagemodel} built upon Qwen-LM with a visual receptor and 3-stage training pipeline, demonstrating strong performance on visual grounding and OCR tasks. Its successor, Qwen2-VL, introduced the Naive Dynamic Resolution mechanism\cite{wang2024qwen2vlenhancingvisionlanguagemodels} for handling variable image resolutions and Multimodal Rotary Position Embedding (M-RoPE)\cite{wang2024qwen2vlenhancingvisionlanguagemodels} for effectively fusing positional information across modalities. Qwen2-VL explored scaling laws across model sizes at 2B, 8B, and 72B parameters, achieving performance competitive with proprietary models at the 72B scale. 

InternVL represents another significant branch of VLM research. The InternVL series scaled vision foundation models to 6B parameters and progressively aligned them with LLMs using web-scale image-text data. InternVL 1.5\cite{chen2024fargpt4vclosinggapInternvl1.5} improved upon this foundation with dynamic high-resolution processing supporting up to 4K resolution input and bilingual dataset enhancements for OCR and Chinese language tasks. The most recent iteration, InternVL 2.5\cite{chen2025expandingperformanceboundariesopensourceinternvl2.5}, maintained the core architecture while focusing on training and testing strategy improvements, achieving high performance on multi-discipline reasoning tasks. 

Microsoft's Phi-3\cite{abdin2024phi3technicalreporthighly} series has extended into the vision domain with Phi-3.5-Vision\cite{abdin2024phi3technicalreporthighly}, a relatively compact 4.2B parameter model derived from the Phi-3.5-mini language model. Despite its modest size, Phi-3.5-Vision demonstrates strong reasoning capabilities and handles both single and multi-image inputs effectively. 

\subsection{Efficiency-Focused Approaches}
A growing trend in VLM research focuses on developing efficient models that maintain high performance while reducing computational requirements. SmolVLM\cite{smolvlm} represents this direction with its 2B parameter model designed for commercial use and local deployment. These models leverage open training pipelines and datasets like Cauldron and Docmatix, demonstrating that smaller models can still achieve practical utility. 

Similarly, Molmo\cite{deitke2024molmopixmoopenweights} introduced a family of VLMs built from scratch without distillation from proprietary models. Their approach combined careful modeling choices with high-quality original created PixMo dataset, including detailed image captions and innovative 2D pointing data. Despite focusing on open development principles, their models achieved competitive performance with larger models. 

Idefics3-8B\cite{laurençon2024buildingbetterunderstandingvisionlanguage} exemplifies efficient VLM development through straightforward training pipelines and exclusive use of open datasets. The creation of Docmatix—a dataset 240 times larger than previously available document understanding resources—contributed significantly to its document processing capabilities. 

\subsection{Document Understanding and OCR Capabilities }
Document understanding and OCR capabilities have become essential benchmarks for evaluating VLM performance. Several models have made notable progress in this domain. InternVL 1.5\cite{chen2024fargpt4vclosinggapInternvl1.5} incorporated high-quality datasets covering document images with bilingual annotations, significantly enhancing OCR-related task performance. Qwen-VL\cite{bai2023qwenvlversatilevisionlanguagemodel} implemented text-reading ability by aligning image-caption-box tuples, while Qwen2-VL's dynamic resolution approach improved document processing capabilities. 

The development of the Docmatix dataset by Idefics3\cite{laurençon2024buildingbetterunderstandingvisionlanguage} marks a significant milestone in advancing document understanding, providing training resources at unprecedented scale. This development has raised the baseline for document processing capabilities in modern VLMs.  

\subsection{Training Strategies and Data Efficiency}
Training methodologies have diversified across VLM development. The Qwen-VL\cite{bai2023qwenvlversatilevisionlanguagemodel} series employed a 3-stage training pipeline with multilingual multimodal cleaned corpus, while Molmo\cite{deitke2024molmopixmoopenweights} emphasized dataset quality over quantity with their carefully curated PixMo datasets. InternVL explored continuous learning strategies for large-scale vision foundation models and high-quality bilingual dataset curation. 

While many approaches have focused on scaling both model size and training data volume as seen with Qwen2-VL's 72B parameter model and InternVL's extensive data collection, our work with Shakti VLM contributes to this landscape by introducing architectural innovations specifically designed to improve data efficiency. Through adopting QK-Normalization\cite{henry2020querykeynormalizationtransformers} for attention stability, hybrid normalization techniques, and enhanced positional encoding, Shakti-VLM models achieve competitive performance despite using fewer training tokens than comparable models. This focus on efficiency through architectural design rather than sheer data volume positions Shakti-VLM as a practical solution for enterprise-scale multimodal tasks. 

\section{Architecture of Shakti-VLM }
The Shakti-VLM-1B and Shakti-VLM-4B models are designed to provide multi-modal understanding through an efficient combination of vision encoding, projection layers, and textual decoding. Both models leverage dynamic patch sizes and hybrid normalization techniques to enhance stability and scalability, along with RoPE\cite{su2023roformerenhancedtransformerrotaryrope} with 2D positional bias and hybrid activation functions ensure improved visual feature extraction. The architecture is designed for tasks like OCR, visual reasoning, and contextual understanding, with decoders optimized for seamless integration of visual and textual modalities. 

\subsection{Vision Encoder}
The vision encoder of the Shakti-VLM-1B model is instantiated upon the Vision Encoder\cite{dosovitskiy2021imageworth16x16wordsvit}, comprising 36 layers, a hidden dimensionality of 1536, and 16 attention heads, optimized for high-resolution visual processing across multiple tasks, including Optical Character Recognition (OCR), fine-grained visual understanding, and image summarization. The encoder incorporates a dynamic patch size mechanism, adaptable within the range of 14×14 at 224px to 32×32 at 1024px resolutions, ensuring robust scalability across varying input image resolutions, which is imperative for precise text recognition and the extraction of intricate visual details. 

The Shakti-VLM-4B model extends this architecture with an expanded vibackbone, comprising 48 layers, a hidden dimensionality of 1920, and 24 attention heads, thereby augmenting its capacity for advanced visual reasoning and scene interpretation. Its dynamic patch size mechanism (ranging from 14×14 to 32×32) ensures high-resolution adaptability. 

Our innovative approach to the Shakti-VLM model design incorporates several key optimizations to enhance stability, efficiency, and precision in multi-modal learning. 

To stabilize attention mechanisms, we utilized the QK-Norm\cite{henry2020querykeynormalizationtransformers}, which applies RMS\cite{zhang2019rootmeansquarelayer} normalization specifically to query and key vectors in attention layers. This rare optimization prevents gradient vanishing/explosion, ensuring robust attention score computations even in deeper architectures. 

We employ a hybrid normalization strategy to further promote convergence stability. Pre-LayerNorm is applied to the initial layers, while RMSNorm\cite{zhang2019rootmeansquarelayer} governs the remaining layers, facilitating smoother optimization. Specifically, in Shakti-VLM-1B, Pre-LayerNorm is utilized for the first 12 layers, with RMSNorm\cite{zhang2019rootmeansquarelayer} applied to the next 24 layers. In contrast, Shakti-VLM-4B implements Pre-LayerNorm across the first 18 layers, followed by RMSNorm\cite{zhang2019rootmeansquarelayer} for the subsequent 30 layers. 

For enhanced spatial encoding and improved object localization, we augment Rotary Position Embedding (RoPE)\cite{su2023roformerenhancedtransformerrotaryrope} with a 2D absolute positional bias, strengthening scene comprehension. Additionally, our use of SiLU\cite{elfwing2017sigmoidweightedlinearunitsneuralsilu} and SwiGLU\cite{shazeer2020gluvariantsimprovetransformerswiglu} activation functions ensures smoother gradient propagation, fostering fine-grained visual feature extraction essential for high-level visual understanding. 

The Shakti-VLM-1B model is trained on 487 billion tokens, equipping it with superior generalization capabilities across diverse visual-language tasks, ensuring high fidelity in contextual reasoning and information retrieval. In contrast, Shakti-VLM-4B is trained on a dataset encompassing 782 billion tokens, fortifying its capacity for multi-modal alignment and high-precision reasoning in image summarization, contextual reasoning, and visual question answering. 

\subsection{Projection Layer}
A projection layer is employed in both models to transform visual features into visual tokens, ensuring their seamless integration with textual inputs. This transformation facilitates robust multi-modal representation learning, enhancing the model’s ability to align and process information across both modalities. The projected visual tokens are then concatenated with text embeddings and fed into the decoder. 

\subsection{Decoder}
The decoder component of the Visual Language Model facilitates the seamless integration of visual and textual modalities for comprehensive multi-modal understanding. The visual representations, processed through a projection layer and encoded as visual tokens, are concatenated with textual inputs and subsequently fed into the decoder. 

The Shakti-VLM-1B employs the Shakti-500M model as its decoder, whereas the Shakti-VLM-4B integrates the Shakti-2.5B\cite{shakhadri2024shakti25billionparameter} model, both of which are optimized for multi-modal alignment and generative reasoning. Utilizing a three-stage training pipeline, the decoders effectively synchronize visual and textual embeddings, facilitating superior performance across tasks such as OCR, image summarization, visual reasoning, and contextual understanding. This architectural design ensures high-precision multi-modal task execution with enhanced efficiency and accuracy.

\section{Training Details }
The training framework for the Shakti-VLM-1B and Shakti-VLM-4B Visual Language Models (VLMs) is divided into three distinct stages: Pre-training Stage 1, Pre-training Stage 2, and Fine-tuning Stage 3. Each stage employs tailored training configurations to incrementally improve the models’ multi-modal comprehension and alignment capabilities. The training parameters shared between the models are summarized in Table \ref{table:shakti-vlm-training-parameters}, while Table \ref{table:shakti-vlm-learning-rates} outlines the learning rate settings for each stage of training. 

\subsection{Pre-training Stage 1}
 This stage focuses exclusively on training the decoder while the encoder remains frozen. The primary objective is to extend the decoder’s context length, thus enhancing its capacity for processing and understanding extended text sequences. For the Shakti-VLM-1B model, the Shakti 500M decoder is optimized to handle sequences of up to 16,384 tokens, whereas the Shakti 2.5B decoder in the Shakti-VLM-4B model is trained to accommodate sequences of up to 32,768 tokens. A cosine learning rate scheduler is employed, with an initial learning rate of 3e-4 for the Shakti-VLM-1B model and 2e-4 for the 4B model. Gradient accumulation steps are set to 2 to ensure efficient learning over large sequence lengths, and rotary position embeddings with dynamic scaling are utilized to handle the extended context lengths. This stage is instrumental in establishing a robust language modeling foundation, enhancing the models’ contextual retention and language comprehension over extended sequences. By isolating the training to the decoder, the models develop refined language generation capabilities before integrating multi-modal inputs in subsequent stages.

\subsection{Pre-training Stage 2 }
 In this stage, the Multi-Layer Perceptron (MLP)\cite{7238334mlp} projector is initialized to bridge the visual and language representations. The primary focus is on training the vision encoder and projection layers to align visual and textual embeddings within the decoder’s input space. The decoder is kept frozen during this stage to prioritize cross-modal alignment. 

For the Shakti-VLM-1B model, the training configuration includes a sequence length of 16,384 tokens, an image size of 448x448, and dynamic resizing to enhance robustness across varied visual inputs. The learning rate is set at 2e-5 with a cosine learning rate schedule.In contrast to the Shakti-VLM-4B model, a longer sequence length of 32,768 tokens is used, along with the same image size and dynamic resizing setup. The learning rate is set to 4e-5, also using a cosine learning rate schedule. 

This stage is pivotal in ensuring effective alignment of visual and textual representations, providing a robust foundation for fine-tuning the models on complex downstream tasks. 

\subsection{Fine-tuning Stage 3 }
 The final stage involves fine-tuning all three components: the encoder, projection layer, and decoder. The models are exposed to a diverse set of image-text datasets to enhance their performance across a range of vision-language tasks. Both the Shakti-VLM-1B and Shakti-VLM-4B models use a learning rate of 4e-5 with a cosine learning rate scheduler and a weight decay of 0.01 to ensure regularization and mitigate overfitting. The image size is maintained at 448x448, with dynamic resizing enabled to ensure adaptability to varied visual contexts. 

This stage is crucial for aligning the visual and textual embeddings across multiple tasks, including Document Visual Question Answering (VQA), Visual Question Answering, and Multimodal Reasoning. Additionally, instruction tuning and in-context instruction tuning are employed to enhance the models' responsiveness to diverse prompts and complex instructions. Reinforcement Learning from Human Feedback (RLHF)\cite{rlhf} is leveraged to refine vision-language outputs based on human preferences, further improving response quality and contextual coherence. Direct Preference Optimization (DPO)\cite{dpo} is integrated to enhance model performance on specialized tasks, ensuring greater adaptability across various real-world applications. Collectively, this comprehensive training process ensures the models’ capacity to understand, reason, and accurately respond to complex multi-modal prompts with high precision and flexibility. 

\subsection{Training Loss Analysis and Convergence}
\begin{figure}[h]
    \centering
    \includegraphics[width=0.5\textwidth]{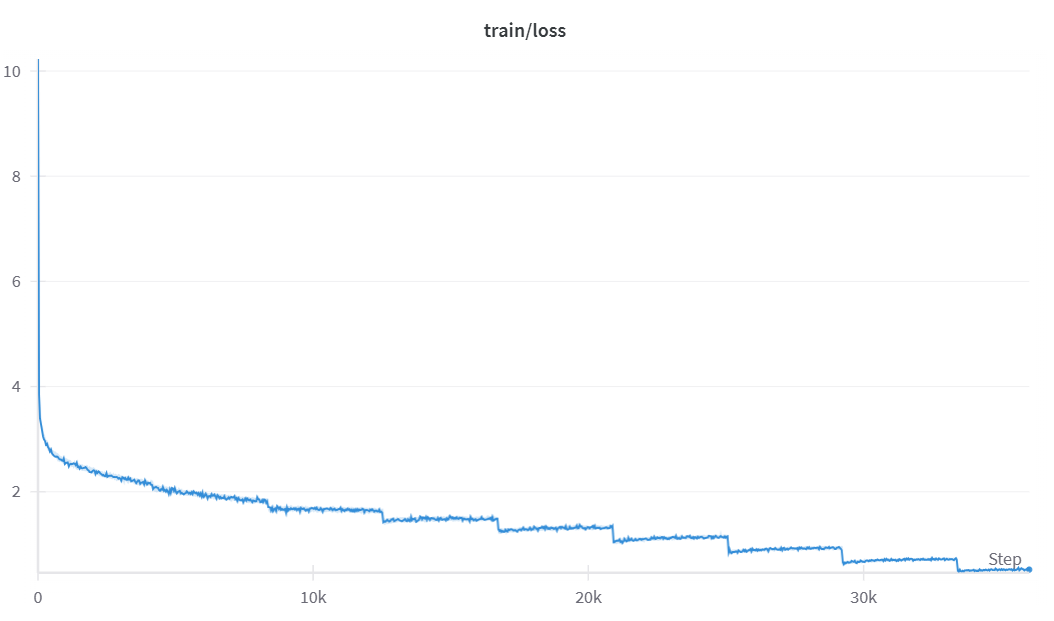}
    \caption{Training Loss Curve for Shakti-VLM-1B:The graph shows the loss reduction from around 10 to 1 over 35k steps, with stepwise drops likely due to scheduled learning rate adjustments, reflecting a more complex training trajectory.}
    \label{fig:1B-trainloss}
\end{figure}

\begin{figure}[h]
    \centering
    \includegraphics[width=0.5\textwidth]{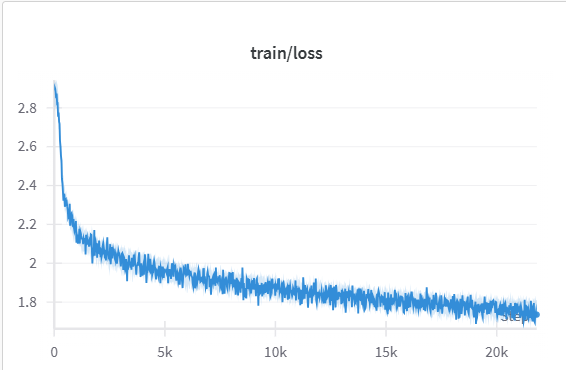}
    \caption{Training Loss Curve for Shakti-VLM-4B – The graph illustrates the steady decline in training loss from approximately 2.8 to 1.8 over 20k steps, indicating stable convergence and effective optimization.}
    \vspace{0.5cm}
    \label{fig:4B-trainloss}
\end{figure}

The training process of Shakti-VLM-1B and Shakti-VLM-4B was analyzed through loss graphs\ref{fig:1B-trainloss} \ref{fig:4B-trainloss}, revealing their convergence behavior and stability. Shakti-VLM-1B began with a high initial loss of approximately 10, which steadily decreased to around 1 over 35k steps, with noticeable stepwise drops likely due to scheduled learning rate adjustments, indicating a more complex optimization trajectory. In contrast, Shakti-VLM-4B started with a training loss of approximately 2.8, gradually reducing to around 1.8 after 20k steps, demonstrating smooth and stable convergence. The models were trained on 8×A100 (40GB) GPUs, processing 487 billion tokens over 12 days for the 1B model and 782 billion tokens over 23 days for the 4B model. Both models achieved effective optimization, with Shakti-VLM-1B experiencing more dynamic loss variations compared to the steadier convergence of Shakti-VLM-4B.

\begin{table}[h]
    \centering
     \renewcommand{\arraystretch}{1.3}
    \begin{tabular}{|l|c|c|}
        \hline
        \textbf{Parameter}        & \textbf{Shakti-VLM-1B} & \textbf{Shakti-4B} \\
        \hline
        LR Scheduler Type         & Cosine             & Cosine             \\
        \hline
        Max\_seq\_len             & 16384              & 32768              \\
        \hline
        Rope\_theta               & 125000             & 500000             \\
        \hline
        Image Size                & 448                & 448                \\
        \hline
        Dynamic Size              & True               & True               \\
        \hline
    \end{tabular}
    \vspace{0.5cm}
    \caption{This table presents the common training parameters for the Shakti-Shakti-VLM-1B and Shakti-4B models, focusing on key settings shared across both models during the training process.}
    \label{table:shakti-vlm-training-parameters}
\end{table}

\begin{table}[h]
    \centering
    \renewcommand{\arraystretch}{1.3}
    \begin{tabular}{|l|c|c|}
        \hline
        \textbf{Training Stages}   & \textbf{Shakti-VLM-1B} & \textbf{Shakti-4B} \\
        \hline
        Pre-Training Stage 1       & 3e-4               & 2e-5               \\
        \hline
        Pre-Training Stage 2       & 2e-5               & 4e-5               \\
        \hline
        Fine-Tuning Stage 3        & 4e-5               & 4e-5               \\
        \hline
    \end{tabular}
    \vspace{1cm}
    \caption{This table outlines the learning rates employed for Shakti-VLM-1B and Shakti-4B models during different pre-training and fine-tuning stages, providing insights into the specific learning rates used to optimize each stage of the training process.}
    \label{table:shakti-vlm-learning-rates}
\end{table}


\section{Dataset Details }
The training of both the Shakti-VLM-1B and Shakti-VLM-4B Visual Language Models leverages a diverse set of datasets spanning various tasks, with the objective of advancing the models' multi-modal comprehension and performance. By exposing the models to a broad spectrum of data modalities, both models acquire robust capabilities for visual-textual alignment, thus facilitating proficiency in complex vision-language tasks. Table \ref{table:Shakti-vlm-dataset}
provides a comprehensive summary of the datasets utilized throughout the three distinct training stages for both the Shakti-VLM-1B and Shakti-VLM-4B models, with particular emphasis on the supplementary datasets used exclusively for the Shakti-VLM-4B model to enhance its performance in visual reasoning, document analysis, and specialized vision-language tasks. 

Text-Only Data: In the initial stage, text-only datasets are employed to establish a solid foundation for language comprehension and increase the context length of the decoder. The Dolma (Books subset)\cite{soldaini2024dolmaopencorpustrillion} is utilized to enhance the models' general language modeling capabilities, while The Stack\cite{Kocetkov2022TheStack} is incorporated to improve code understanding. The FineWeb-Edu-dedup\cite{penedo2024thefineweb} dataset contributes to strengthening the models' overall language understanding. This foundational stage enables the models to process and respond to text-based inputs with a high degree of accuracy and coherence. 

Image and text captioning datasets play a crucial role in aligning visual inputs with their corresponding text descriptions. LAION-400M\cite{schuhmann2022laion5bopenlargescaledataset} and LAION COCO\cite{laioncoco} provide a broad collection of image-caption pairs for general captioning tasks. COCOCaption\cite{chen2015microsoftcococaptionsdata} and TextCaptions\cite{sidorov2019textcaps} are employed to further improve caption generation. These datasets enable the models to produce highly accurate and contextually relevant captions for a wide range of images, improving image-description alignment. 

Document analysis datasets enhance the models' capabilities in processing and understanding structured and semi-structured document images. PDFA\cite{pixparse_pdfa_eng_wds} is utilized for document layout understanding, focusing on the structural relationships in document formatting. DocVQA\cite{mathew2021docvqa} and Docmatrix\cite{laurençon2024buildingbetterunderstandingvisionlanguage} are used for document visual question answering, training the models to answer questions based on document content accurately. These datasets prepare the models for tasks involving complex document processing and information extraction. 

Visual question answering (VQA) datasets are designed to strengthen the models' reasoning and answering capabilities for visual inputs. Datasets such as Visual-7W\cite{zhu2016visual7wgroundedquestionanswering} and OCR-VQA\cite{8978122ocrvqa} focus on answering questions derived from visual elements in both images and text-based documents. LLaVA-CoT-100k\cite{xu2024llavacotletvisionlanguage} and DataComp\cite{gadre2023datacompsearchgenerationmultimodal} provide additional training for context-based VQA tasks. This training equips the models to handle real-world visual and textual queries with precision and relevance. 

Instruction tuning and fine-tuning datasets are used to improve model adaptability for specialized and dynamic tasks. Leopard-instruct is employed for instruction tuning, allowing the models to interpret and follow diverse task instructions. MIMIC-IT\cite{li2023mimicitmultimodalincontextinstruction} supports in-context instruction tuning to adapt to various prompts and queries. The cauldron and rlaif-v-formatted\cite{yu2024rlaifv} are fine-tuning datasets that enhance the models' performance on downstream vision-language tasks. RLAIF-V-Dataset\cite{yu2024rlaifv} is used for reinforcement learning from human feedback (RLHF)\cite{rlhf}, aligning the models' outputs with human preferences. 

Specialized reasoning and multimodal tasks are supported by datasets such as ScienceQA\cite{lu2022learnscienceqa}, which focuses on science-based question answering with multi-modal inputs. This dataset improves the models' ability to reason through complex visual and language information in a logical and coherent manner. 

By leveraging this diverse set of datasets across different task categories, the Shakti-VLM-1B and Shakti-VLM-4B model develop comprehensive multi-modal alignment capabilities. This diverse training pipeline enables the models to excel at tasks such as image captioning, document analysis, visual reasoning, and visual question answering with high accuracy, adaptability, and contextual understanding across multiple domains.

\begin{table}
\captionsetup{position=bottom, skip=0.5cm}

\centering
\renewcommand{\arraystretch}{1.8} 
\setlength{\tabcolsep}{4pt} 
\resizebox{\textwidth}{!}{%
\begin{tabular}{|p{2.5cm}|p{4cm}|p{4cm}|p{4cm}|}
\hline
\rule{0pt}{3.5ex} & \multicolumn{2}{c|}{\textbf{Pre-Training}} & \multicolumn{1}{c|}{\textbf{Fine-Tuning}} \\
\cline{2-4}
\rule{0pt}{3.5ex} & \multicolumn{1}{c|}{Stage 1} & \multicolumn{1}{c|}{Stage 2} & \multicolumn{1}{c|}{Stage 3} \\
\hline
\rule{0pt}{4.5ex}
Used for both Shakti-VLM-1B and Shakti-VLM-4B & 
\begin{minipage}[t]{4cm}
\begin{itemize}
  \item \textit{Dolma (Books subset)}
  \item \textit{The Stack, FineWeb-Edu-dedup}
\end{itemize}
\end{minipage} & 
\begin{minipage}[t]{4cm}
\begin{itemize}
  \item \textit{OBELICS}
  \item \textit{PDFA}
  \item \textit{LAION-400M}
\end{itemize}
\end{minipage} & 
\begin{minipage}[t]{4cm}
\begin{itemize}
  \item \textit{PDFA}
  \item \textit{Docmatrix}
  \item \textit{Leopard-instruct}
  \item \textit{MIMIC-IT}
  \\
\end{itemize}
\end{minipage} \\
\hline

\rule{0pt}{4.5ex}
Used for only Shakti-VLM-4B & 
\begin{minipage}[t]{4cm}
\begin{itemize}
  \item \textit{Dolma (Books subset)}
  \item \textit{The Stack, FineWeb-Edu-dedup}
\end{itemize}
\end{minipage} & 
\begin{minipage}[t]{4cm}
\begin{itemize}
  \item \textit{LAION COCO}
  \item \textit{COYO}
  \item \textit{DocVQA}
  \item \textit{TextCaptions}
  \item \textit{Visual-7W}
  \item \textit{OCR-VQA}
  \item \textit{DataComp}
  \item \textit{COCOCaption}
  \\
\end{itemize}
\end{minipage} & 
\begin{minipage}[t]{4cm}
\begin{itemize}
  \item \textit{ScienceQA}
  \item \textit{RLAIF-V-Dataset}
  \item \textit{LLaVA-CoT-100k}
  \item \textit{the\_cauldron}
  \item \textit{rlaif-v\_formatted}
\end{itemize}
\end{minipage} \\
\hline
\end{tabular}%
}
\caption{Datasets used across different training stages for both Shakti-VLM-1B and 4B models, highlighting additional datasets utilized exclusively for the 4B model to enhance multi-modal performance.}
\label{table:Shakti-vlm-dataset}
\end{table}

\section{Evaluation and Results}
We evaluated our Shakti Vision Language Models across a diverse set of multimodal benchmarks, comparing their performance with contemporary models within similar parameter ranges. For the Shakti-VLM-1B model, comparisons were made against several popular VLM models in the 1B to 3B parameter range. For the Shakti-VLM-4B model, we compared leading models in the 4B to 8B parameter range. 

Our comprehensive benchmarking approach encompasses a wide spectrum of multimodal understanding tasks, including document understanding, chart interpretation, mathematical reasoning, and general vision-language capabilities. The benchmark suite was carefully selected to evaluate multiple dimensions of model performance: OCR capabilities, visual reasoning, complex multimodal understanding, mathematical reasoning, and practical real-world applications. 

The benchmark datasets represent diverse challenges in the multimodal domain, with varying degrees of complexity. This diversity allows us to thoroughly assess each model's generalization capabilities across different task types, data distributions, and reasoning requirements. The benchmark results for Shakti-VLM-1B and Shakti-VLM-4B models are showcased in the table \ref{table:benchmark-shakti-vlm-1B} and table \ref{table:benchmark-shakti-VLM-4B} respectively along with the comparison models. 

\begin{table}

\centering
\renewcommand{\arraystretch}{2} 
\setlength{\tabcolsep}{3.8pt} 
\resizebox{\textwidth}{!}{%
\begin{tabular}{|l|c|c|c|c|c|c|c|c|}
\hline
\textbf{Benchmarks} & \textbf{Shakti-VLM-1B} & \textbf{MolmoE-1B} & \textbf{InternVL2-1B} & \textbf{SmolVLM-2.25B} & \textbf{MiniCPM-V-2.0-2.8B} & \textbf{Qwen-2VL-2B} & \textbf{InternVL2-2B} & \textbf{Qwen-2.5VL-3B} \\
\hline
MMMUval & \underline{42.5} & 34.9 & 36.7 & 38.8 & 38.2 & 41.1 & 36.3 & \textbf{53.1} \\
\hline
DocVQAtest & 87.96 & 77.7 & 81.7 & 81.6 & 71.9 & \underline{90.1} & 86.9 & \textbf{93.9} \\
\hline
InfoVQAtest & 56.8 & 53.9 & 50.9 & 43.5 & 49.1 & \underline{65.5} & 58.9 & \textbf{77.1} \\
\hline
ChartQAtest & \textbf{79.56} & \underline{78} & 72.9 & 62.2 & 70.1 & 73.5 & 76.2 & - \\
\hline
TextVQAval & \textbf{80.75} & 78.8 & 70.5 & 72.7 & 74.1 & \underline{79.7} & 73.4 & 79.3 \\
\hline
OCRBench & \textbf{798} & 684 & 754 & 701 & 605 & \underline{794} & 781 & - \\
\hline
MMEsum & \textbf{1910.62} & 1782.2 & 1794.4 & 1801.9 & 1808.6 & 1872 & \underline{1876.8} & - \\
\hline
MMStar & \underline{50.13} & 40.2 & 39.4 & 42.1 & 46.8 & 48 & 49.8 & \textbf{55.9} \\
\hline
MMMU Pro val & \underline{24.73} & - & - & - & - & - & - & \textbf{31.6} \\
\hline
VQA v2val & \underline{76.28} & \textbf{83.9} & 69.5 & 58.2 & 66.4 & 71.2 & 67.6 & - \\
\hline
Ai2d & 77.29 & \textbf{86.4} & 64.1 & 61.9 & 55.4 & - & - & \underline{81.5} \\
\hline
RealworldQA & \textbf{64.82} & 60.4 & 50.3 & - & - & \underline{62.9} & 57.3 & - \\
\hline
MathVista (testmini) & \underline{46.2} & 34 & 37.7 & 44.6 & 38.7 & - & - & \textbf{62.3} \\
\hline
MMT-Bench (test) & \textbf{57.4} & 52.1 & 48.9 & - & - & \underline{54.5} & - & - \\
\hline
MMVet & \textbf{44.9} & - & \underline{32.7} & - & - & - & - & - \\
\hline
HallusionBench & \textbf{40.07} & - & \underline{34} & - & - & - & - & - \\
\hline
MMBench (test) & \underline{42.4} & - & - & - & - & - & - & \textbf{77.6} \\
\hline
MathVision & 17.03 & - & 12.2 & - & - & \underline{19.7} & 15.8 & \textbf{21.2} \\
\hline
MathVerse & 19 & - & 18.4 & - & - & \underline{21} & \textbf{25.3} & - \\
\hline
Olympaid Bench & \textbf{0.9} & - & 0.3 & - & - & - & \underline{0.4} & - \\
\hline
BLINK & 39.9 & - & 38.6 & - & - & \textbf{44.4} & \underline{43.8} & - \\
\hline
MTVQA  & \underline{13.2} & - & 12.6 & - & - & \textbf{20} & 10.9 & - \\
\hline
\end{tabular}%
}
\vspace{0.5cm}
\caption{ Benchmark Performance Comparison of Shakti-VLM-1B model against other VLM models in the parameter range of 1B to 3B parameters.}

\label{table:benchmark-shakti-vlm-1B}
\end{table}

\subsection{Results for Shakti-VLM-1B}
\subsubsection{Performance Highlights}
Shakti-VLM-1B demonstrates exceptional performance across multiple benchmarks despite its compact size. The model achieves high performance in several key areas: 
\begin{itemize}
    \item \textbf{MMMU (Multimodal Massive Multitask Understanding):} Shakti-VLM-1B achieves 42.5\% on the validation set, surpassing all comparison models of same parameter and compitative to the latest Qwen-2.5VL 3B\cite{bai2025qwen25vltechnicalreport} model.

    \item \textbf{Document and Text Understanding:} Strong performance on DocVQA , TextVQA, and OCRBench demonstrates the model's robust text recognition and document understanding capabilities. 

    \item \textbf{Chart Understanding:} Leading performance on ChartQA  indicates superior ability to interpret and reason about visual data representations.

    \item \textbf{General Multimodal Evaluation:} Shakti-VLM-1B achieves the highest score on MME  with 1910.62 points and MMStar with 50.13\%, showcasing its balanced capabilities across diverse multimodal tasks.

    \item \textbf{Mathematical Reasoning:} Strongest performance on MathVista among models in its size class, demonstrating advanced visual mathematical reasoning capabilities and competitive to the latest model.
\end{itemize}

\subsubsection{Comparative Analysis }
When compared to models of similar or larger sizes, Shakti-VLM-1B shows several notable strengths:

\begin{itemize}
    \item Balanced Performance: While some comparison models excel in specific domains, Shakti-VLM-1B maintains high performance across a broader spectrum of tasks, suggesting better generalization capabilities. 

    \item Shakti-VLM-1B frequently outperforms models with significantly more parameters, such as SmolVLM-2.25B\cite{smolvlm} and MiniCPM-V-2.0-2.8B\cite{yao2024minicpm26}, highlighting the efficiency of its architecture and training methodology. 

    \item Strong Document and Diagram Understanding: The model demonstrates particular strength in tasks requiring joint reasoning over text and visual elements, as evidenced by its leading performance on ChartQA and strong results on DocVQA and TextVQA. 

    \item Mathematical Reasoning: Strong performance on MathVista demonstrates advanced visual mathematical reasoning capabilities, significantly outperforming MolmoE-1B\cite{deitke2024molmopixmoopenweights}  and MiniCPM-V-2.0-2.8B\cite{yao2024minicpm26}. 
\end{itemize}

\subsection{Results for Shakti-VLM-4B}
\subsubsection{Performance Highlights}

Shakti-VLM-4B demonstrates substantial improvements over its Shakti-VLM-1B counterpart and achieves excellent results across numerous benchmarks: 

\begin{itemize}
    \item \textbf{Comprehensive Understanding:} Exceptional performance on MMMU (59.78\%), significantly outperforming all comparison models, indicating superior capabilities in complex multimodal reasoning tasks.

    \item \textbf{Document Intelligence:} The results on DocVQA, TextVQA and InfoVQA demonstrates the model capability in the document understanding.

    \item \textbf{Visual Reasoning:} The performance on ChartQA, MMStar , and MMVet  showcases the model's advanced visual reasoning abilities.

    \item \textbf{Real-world Application: } Highest scores on RealworldQA suggest superior practical applicability in everyday scenarios.
\end{itemize}


\begin{table}
\captionsetup{position=bottom, skip=0.5cm}
\centering
\renewcommand{\arraystretch}{1.7} 
\setlength{\tabcolsep}{3.8pt} 
\resizebox{\textwidth}{!}{%
\begin{tabular}{|l|c|c|c|c|c|c|}
\hline
\textbf{Benchmarks} & \textbf{Shakti-4B} & \textbf{InternVL2-4B} & \textbf{Phi-3-Vision-4B} & \textbf{MiniCPM-V-2.6-8B} & \textbf{Qwen2VL-7B} & \textbf{Qwen2.5VL-7B} \\
\hline
MMMUval & \textbf{59.78} & 47.9 & 46.1 & 49.8 & 54.1 & \underline{58.6} \\
\hline
DocVQAtest & 92.92 & 89.2 & - & 90.8 & \underline{94.5} & \textbf{95.7} \\
\hline
InfoVQAtest & \underline{77.3} & 67.0 & - & - & 76.5 & \textbf{ 82.6} \\
\hline
ChartQAtest & \underline{85.28} & 81.5 & 81.4 & - & 83.0 & \textbf{87.3} \\
\hline
TextVQAval & \textbf{85.56} & 74.4 & 70.9 & 80.1 & 84.3 & \underline{84.9} \\
\hline
OCRBench & 849 & 788 & 639 & \underline{852} & 845 & \textbf{864} \\
\hline
MMEsum & \underline{2340.99} & 2064.1 & 1508.0 & \textbf{2348.4} & 2326.8 & - \\
\hline
MMStar & \underline{62.33} & - & - & 57.5 & 60.7 & \textbf{63.9} \\
\hline
MMMU Pro val & \underline{37.47} & - & - & - &- & \textbf{41} \\
\hline
VQA v2val & 78.78 & - & - & - & - & - \\
\hline
Ai2d & \textbf{83.83} & \underline{78.9} & 76.7 & - & - & - \\
\hline
RealworldQA & \textbf{71.18} & 60.7 & 58.8 & - &\underline{70.1} & - \\
\hline
MathVista (testmini) & 48.5 & 58.6 & 44.5 & \underline{60.6} & 58.2 & \textbf{68.2} \\
\hline
MMT-Bench (test) & \textbf{66.26} & - & - & - & 63.7 &\underline{63.6} \\
\hline
MMVet &\underline{62.3} & 55.7 & 44.1 & 60 & 62.0 & \textbf{67.1} \\
\hline
HallusionBench & 47.9 & 41.9 & 39 & 48.1 & \underline{50.6} & \textbf{52.9} \\
\hline
MMBench (test) & 81.7 & 78.6 & 73.6 & - & \underline{83.0} & \textbf{82.6} \\
\hline
MathVision & \underline{19.05} & 17.8 & 17.4 & 16.1 & 16.3 & \textbf{25.07} \\
\hline
MathVerse & 28.78 & \underline{32} & 24.1 & 25.7 & \textbf{31.9} & - \\
\hline
Olympaid Bench & \textbf{1.3} & \underline{1.1} & - & - & - & - \\
\hline
BLINK & 50.11 & 46.1 & \textbf{58.3} & 53 &\underline{53.2} & - \\
\hline
MTVQA & \underline{16.02} & 15.3 & - & - & \textbf{25.6} & - \\
\hline
\end{tabular}%
}
\caption{Benchmark Performance Comparison of Shakti-VLM-4B model against other VLM models in the parameter range of 4B to 8B parameters.}
\label{table:benchmark-shakti-VLM-4B}
\end{table}

\subsubsection{Comparative Analysis}
When compared to contemporary models in the 4B to 8B parameter range: 

\begin{itemize}
    \item Consistent Outperformance: Shakti-VLM-4B achieves high performance in most of the benchmarks, showcasing its consistency across a variety of tasks. 

    \item Efficiency vs. Larger Models: Despite having fewer parameters than Qwen2VL-7B\cite{wang2024qwen2vlenhancingvisionlanguagemodels} and MiniCPM-V-2.6-8B\cite{yao2024minicpm26} and latest Qwen-2.5VL 7B\cite{bai2025qwen25vltechnicalreport}, Shakti-VLM-4B achieves comparable and better performance across most benchmarks, highlighting its parameter efficiency. 

    \item Balanced Capabilities: While some models demonstrate strength in specific domains, Shakti-VLM-4B maintains high performance across diverse task types, suggesting more balanced and generalizable capabilities. 

    \item Mathematical and Visual Reasoning: Strong performance on complex reasoning tasks like MathVista, MMVet, and MMT-Bench demonstrates the model's advanced reasoning capabilities. 
\end{itemize}

The comprehensive evaluation results presented above demonstrate that Shakti-VLM models achieve exceptional performance across diverse multimodal tasks, frequently outperforming contemporary models. These findings validate our architectural innovations and training methodology, positioning Shakti as a highly competitive solution for real-world multimodal applications.

\subsection{Qualitative evaluation}

\begin{figure}[h]
    \centering
    \includegraphics[width=0.8\textwidth]{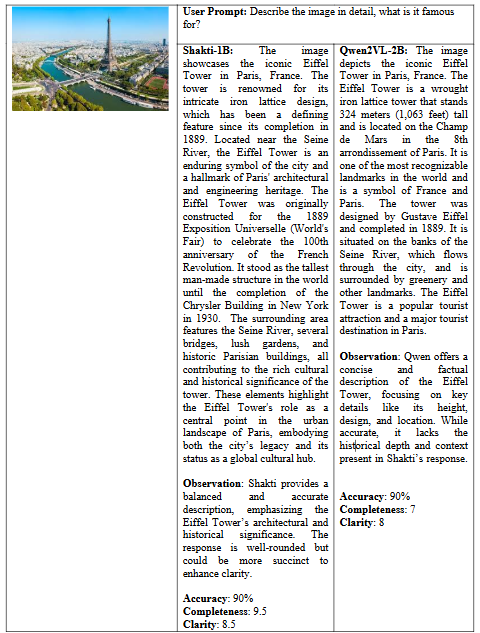}
    \caption{Comparision of Shakti-1B and Qwen2VL-2B Results on different prompts.}
    \label{fig:1B-prompt}
\end{figure}

\begin{figure}[h]
    \centering
    \includegraphics[width=0.8\textwidth]{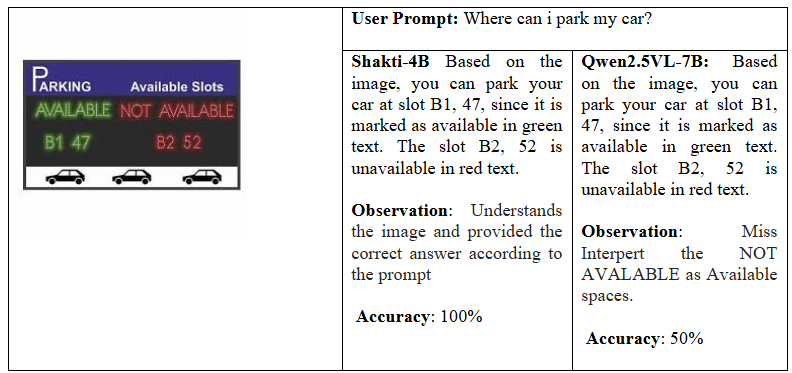}
    \caption{Comparision of Shakti-4B and Qwen2.5VL-7B Results on different prompts.}
    \label{fig:4B-prompt}
\end{figure}

We evaluated Shakti-VLM-1B, Shakti-VLM-4B, Qwen2VL-2B, and Qwen2.5VL-7B across multiple tasks\ref{fig:1B-prompt}, \ref{fig:4B-prompt}, highlighting key performance distinctions. In descriptive tasks, Shakti-VLM-1B demonstrated greater contextual depth, particularly in historical and architectural analyses, whereas Qwen2VL-2B\cite{wang2024qwen2vlenhancingvisionlanguagemodels} prioritized concise factual reporting. For handwritten text extraction, both models exhibited high accuracy with minor spelling errors; however, Shakti-VLM-1B replicated source errors rather than correcting them. In multiple-choice question answering, both models accurately identified the correct responses, reflecting strong factual comprehension.

Summarization tasks revealed notable differences, with Shakti-VLM-4B effectively capturing broader themes, while Qwen2.5VL-7B\cite{bai2025qwen25vltechnicalreport} introduced inaccuracies related to image captioning and datasets. In code generation, Shakti-VLM-4B provided both correct solutions and explanatory context, whereas Qwen2.5VL-7B omitted justifications. Visual reasoning assessments, such as identifying available parking slots, further demonstrated Shakti-4B’s superior interpretative accuracy, as Qwen2.5VL-7B misread indicators.

Overall, Shakti models exhibited stronger contextual depth and reasoning capabilities, while Qwen models, though concise, occasionally introduced interpretative errors. These findings suggest that Shakti models are better suited for complex analytical tasks, whereas Qwen models favor brevity but may require refinement in reasoning-driven applications.

\begin{figure}[h]
    \centering
    \includegraphics[width=0.8\textwidth]{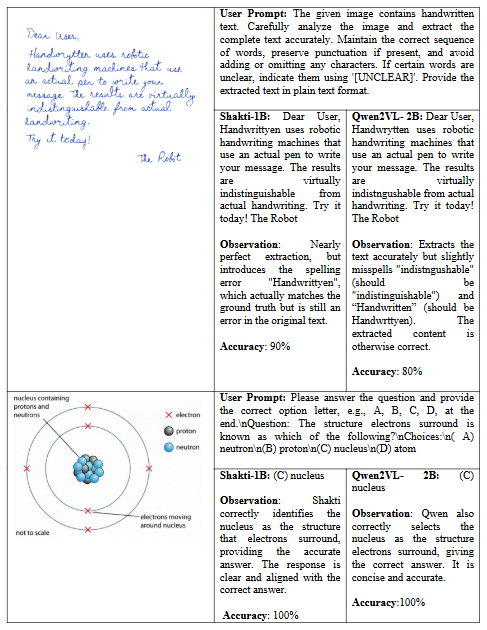}
\end{figure}

\begin{figure}[h]
    \centering
    \includegraphics[width=0.8\textwidth]{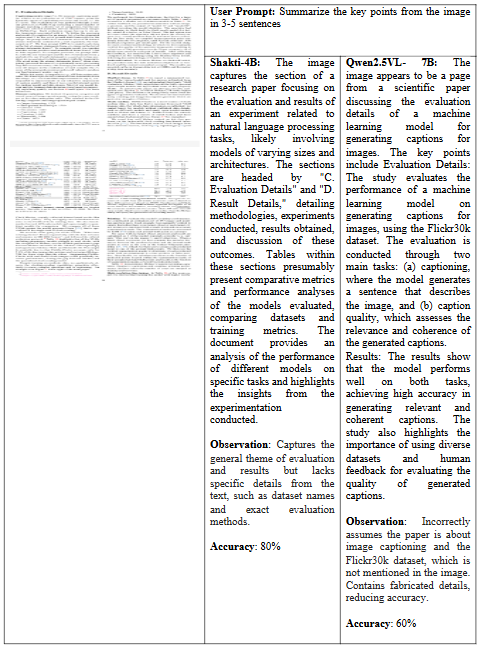}
\end{figure}
\begin{figure}[h]
    \centering
    \includegraphics[width=0.8\textwidth]{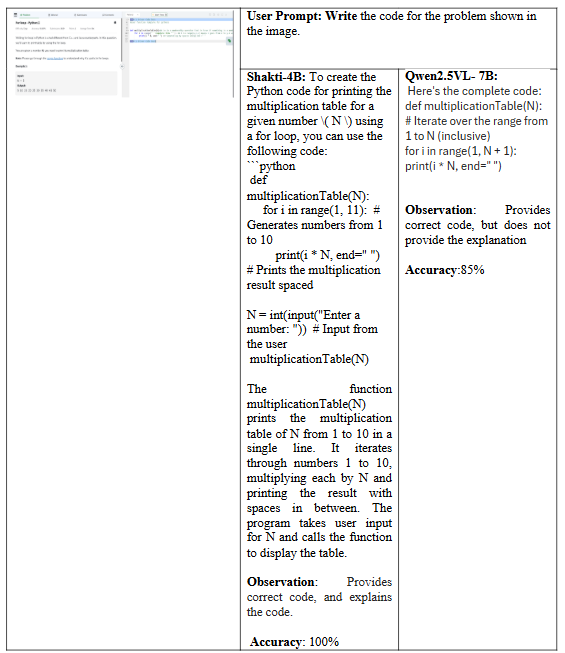}
\end{figure}

\section{Conclusion }
Shakti VLM presents a novel approach to vision-language modeling by emphasizing architectural efficiency and training optimization rather than sheer data volume. By incorporating QK-Normalization\cite{henry2020querykeynormalizationtransformers}, hybrid normalization techniques, and enhanced positional encoding, Shakti-VLM-1B and Shakti-VLM-4B achieve competitive performance on various multimodal tasks such as document understanding, OCR extraction, and general reasoning. Despite using significantly fewer training tokens than comparable models, Shakti models outperform several state-of-the-art alternatives, demonstrating the effectiveness of our three-stage training strategy. These results highlight the potential of intelligent model design in advancing efficient and scalable multimodal AI solutions.

\section{Future Works}
Future work on Shakti VLM presents several promising directions for further exploration. Scaling to larger models beyond 4B parameters could help assess whether the architectural innovations continue to yield efficiency improvements at greater scales. Enhancing data efficiency by exploring additional pretraining strategies, such as curriculum learning and contrastive learning, may reduce reliance on large-scale datasets. Fine-tuning for specialized domains, including medical imaging, legal document analysis, and financial reporting, can further expand its applicability. Optimizing real-time inference speed and efficiency is crucial for deployment in enterprise use cases. Additionally, expanding Shakti’s VLM models capabilities to support multilingual and multimodal inputs, such as audio and video processing, will further enhance its versatility. 

\clearpage

\bibliographystyle{unsrt}  
\bibliography{references}

\begin{thebibliography}{10}

\bibitem{dosovitskiy2021imageworth16x16wordsvit}
Alexey Dosovitskiy, Lucas Beyer, Alexander Kolesnikov, Dirk Weissenborn, Xiaohua Zhai, Thomas Unterthiner, Mostafa Dehghani, Matthias Minderer, Georg Heigold, Sylvain Gelly, Jakob Uszkoreit, and Neil Houlsby.
\newblock An image is worth 16x16 words: Transformers for image recognition at scale, 2021.

\bibitem{wang2024qwen2vlenhancingvisionlanguagemodels}
Peng Wang, Shuai Bai, Sinan Tan, Shijie Wang, Zhihao Fan, Jinze Bai, Keqin Chen, Xuejing Liu, Jialin Wang, Wenbin Ge, Yang Fan, Kai Dang, Mengfei Du, Xuancheng Ren, Rui Men, Dayiheng Liu, Chang Zhou, Jingren Zhou, and Junyang Lin.
\newblock Qwen2-vl: Enhancing vision-language model's perception of the world at any resolution, 2024.

\bibitem{deitke2024molmopixmoopenweights}
Matt Deitke, Christopher Clark, Sangho Lee, Rohun Tripathi, Yue Yang, Jae~Sung Park, Mohammadreza Salehi, Niklas Muennighoff, Kyle Lo, Luca Soldaini, Jiasen Lu, Taira Anderson, Erin Bransom, Kiana Ehsani, Huong Ngo, YenSung Chen, Ajay Patel, Mark Yatskar, Chris Callison-Burch, Andrew Head, Rose Hendrix, Favyen Bastani, Eli VanderBilt, Nathan Lambert, Yvonne Chou, Arnavi Chheda, Jenna Sparks, Sam Skjonsberg, Michael Schmitz, Aaron Sarnat, Byron Bischoff, Pete Walsh, Chris Newell, Piper Wolters, Tanmay Gupta, Kuo-Hao Zeng, Jon Borchardt, Dirk Groeneveld, Crystal Nam, Sophie Lebrecht, Caitlin Wittlif, Carissa Schoenick, Oscar Michel, Ranjay Krishna, Luca Weihs, Noah~A. Smith, Hannaneh Hajishirzi, Ross Girshick, Ali Farhadi, and Aniruddha Kembhavi.
\newblock Molmo and pixmo: Open weights and open data for state-of-the-art vision-language models, 2024.

\bibitem{smolvlm}
Hugging Face.
\newblock Smolvlm:small yet mighty vision language model.
\newblock https://huggingface.co/blog/smolvlm, 2024.
\newblock Accessed: Feb. 22, 2025.

\bibitem{henry2020querykeynormalizationtransformers}
Alex Henry, Prudhvi~Raj Dachapally, Shubham Pawar, and Yuxuan Chen.
\newblock Query-key normalization for transformers, 2020.

\bibitem{rlhf}
Long Ouyang, Jeff Wu, Xu~Jiang, Diogo Almeida, Carroll~L. Wainwright, Pamela Mishkin, Chong Zhang, Sandhini Agarwal, Katarina Slama, Alex Ray, John Schulman, Jacob Hilton, Fraser Kelton, Luke Miller, Maddie Simens, Amanda Askell, Peter Welinder, Paul Christiano, Jan Leike, and Ryan Lowe.
\newblock Training language models to follow instructions with human feedback, 2022.

\bibitem{dpo}
Rafael Rafailov, Archit Sharma, Eric Mitchell, Stefano Ermon, Christopher~D. Manning, and Chelsea Finn.
\newblock Direct preference optimization: Your language model is secretly a reward model, 2024.

\bibitem{yao2024minicpm26}
Yuan Yao, Tianyu Yu, Ao~Zhang, Chongyi Wang, Junbo Cui, Hongji Zhu, Tianchi Cai, Haoyu Li, Weilin Zhao, Zhihui He, et~al.
\newblock Minicpm-v: A gpt-4v level mllm on your phone.
\newblock {\em arXiv preprint arXiv:2408.01800}, 2024.

\bibitem{bai2023qwenvlversatilevisionlanguagemodel}
Jinze Bai, Shuai Bai, Shusheng Yang, Shijie Wang, Sinan Tan, Peng Wang, Junyang Lin, Chang Zhou, and Jingren Zhou.
\newblock Qwen-vl: A versatile vision-language model for understanding, localization, text reading, and beyond, 2023.

\bibitem{chen2024fargpt4vclosinggapInternvl1.5}
Zhe Chen, Weiyun Wang, Hao Tian, Shenglong Ye, Zhangwei Gao, Erfei Cui, Wenwen Tong, Kongzhi Hu, Jiapeng Luo, Zheng Ma, Ji~Ma, Jiaqi Wang, Xiaoyi Dong, Hang Yan, Hewei Guo, Conghui He, Botian Shi, Zhenjiang Jin, Chao Xu, Bin Wang, Xingjian Wei, Wei Li, Wenjian Zhang, Bo~Zhang, Pinlong Cai, Licheng Wen, Xiangchao Yan, Min Dou, Lewei Lu, Xizhou Zhu, Tong Lu, Dahua Lin, Yu~Qiao, Jifeng Dai, and Wenhai Wang.
\newblock How far are we to gpt-4v? closing the gap to commercial multimodal models with open-source suites, 2024.

\bibitem{chen2025expandingperformanceboundariesopensourceinternvl2.5}
Zhe Chen, Weiyun Wang, Yue Cao, Yangzhou Liu, Zhangwei Gao, Erfei Cui, Jinguo Zhu, Shenglong Ye, Hao Tian, Zhaoyang Liu, Lixin Gu, Xuehui Wang, Qingyun Li, Yimin Ren, Zixuan Chen, Jiapeng Luo, Jiahao Wang, Tan Jiang, Bo~Wang, Conghui He, Botian Shi, Xingcheng Zhang, Han Lv, Yi~Wang, Wenqi Shao, Pei Chu, Zhongying Tu, Tong He, Zhiyong Wu, Huipeng Deng, Jiaye Ge, Kai Chen, Kaipeng Zhang, Limin Wang, Min Dou, Lewei Lu, Xizhou Zhu, Tong Lu, Dahua Lin, Yu~Qiao, Jifeng Dai, and Wenhai Wang.
\newblock Expanding performance boundaries of open-source multimodal models with model, data, and test-time scaling, 2025.

\bibitem{abdin2024phi3technicalreporthighly}
Marah Abdin, Jyoti Aneja, Hany Awadalla, Ahmed Awadallah, Ammar~Ahmad Awan, Nguyen Bach, Amit Bahree, Arash Bakhtiari, Jianmin Bao, Harkirat Behl, Alon Benhaim, Misha Bilenko, Johan Bjorck, Sébastien Bubeck, Martin Cai, Qin Cai, Vishrav Chaudhary, Dong Chen, Dongdong Chen, Weizhu Chen, Yen-Chun Chen, Yi-Ling Chen, Hao Cheng, Parul Chopra, Xiyang Dai, Matthew Dixon, Ronen Eldan, Victor Fragoso, Jianfeng Gao, Mei Gao, Min Gao, Amit Garg, Allie~Del Giorno, Abhishek Goswami, Suriya Gunasekar, Emman Haider, Junheng Hao, Russell~J. Hewett, Wenxiang Hu, Jamie Huynh, Dan Iter, Sam~Ade Jacobs, Mojan Javaheripi, Xin Jin, Nikos Karampatziakis, Piero Kauffmann, Mahoud Khademi, Dongwoo Kim, Young~Jin Kim, Lev Kurilenko, James~R. Lee, Yin~Tat Lee, Yuanzhi Li, Yunsheng Li, Chen Liang, Lars Liden, Xihui Lin, Zeqi Lin, Ce~Liu, Liyuan Liu, Mengchen Liu, Weishung Liu, Xiaodong Liu, Chong Luo, Piyush Madan, Ali Mahmoudzadeh, David Majercak, Matt Mazzola, Caio César~Teodoro Mendes, Arindam Mitra, Hardik Modi, Anh Nguyen,
  Brandon Norick, Barun Patra, Daniel Perez-Becker, Thomas Portet, Reid Pryzant, Heyang Qin, Marko Radmilac, Liliang Ren, Gustavo de~Rosa, Corby Rosset, Sambudha Roy, Olatunji Ruwase, Olli Saarikivi, Amin Saied, Adil Salim, Michael Santacroce, Shital Shah, Ning Shang, Hiteshi Sharma, Yelong Shen, Swadheen Shukla, Xia Song, Masahiro Tanaka, Andrea Tupini, Praneetha Vaddamanu, Chunyu Wang, Guanhua Wang, Lijuan Wang, Shuohang Wang, Xin Wang, Yu~Wang, Rachel Ward, Wen Wen, Philipp Witte, Haiping Wu, Xiaoxia Wu, Michael Wyatt, Bin Xiao, Can Xu, Jiahang Xu, Weijian Xu, Jilong Xue, Sonali Yadav, Fan Yang, Jianwei Yang, Yifan Yang, Ziyi Yang, Donghan Yu, Lu~Yuan, Chenruidong Zhang, Cyril Zhang, Jianwen Zhang, Li~Lyna Zhang, Yi~Zhang, Yue Zhang, Yunan Zhang, and Xiren Zhou.
\newblock Phi-3 technical report: A highly capable language model locally on your phone, 2024.

\bibitem{laurençon2024buildingbetterunderstandingvisionlanguage}
Hugo Laurençon, Andrés Marafioti, Victor Sanh, and Léo Tronchon.
\newblock Building and better understanding vision-language models: insights and future directions, 2024.

\bibitem{su2023roformerenhancedtransformerrotaryrope}
Jianlin Su, Yu~Lu, Shengfeng Pan, Ahmed Murtadha, Bo~Wen, and Yunfeng Liu.
\newblock Roformer: Enhanced transformer with rotary position embedding, 2023.

\bibitem{zhang2019rootmeansquarelayer}
Biao Zhang and Rico Sennrich.
\newblock Root mean square layer normalization, 2019.

\bibitem{elfwing2017sigmoidweightedlinearunitsneuralsilu}
Stefan Elfwing, Eiji Uchibe, and Kenji Doya.
\newblock Sigmoid-weighted linear units for neural network function approximation in reinforcement learning, 2017.

\bibitem{shazeer2020gluvariantsimprovetransformerswiglu}
Noam Shazeer.
\newblock Glu variants improve transformer, 2020.

\bibitem{shakhadri2024shakti25billionparameter}
Syed Abdul~Gaffar Shakhadri, Kruthika KR, and Rakshit Aralimatti.
\newblock Shakti: A 2.5 billion parameter small language model optimized for edge ai and low-resource environments, 2024.

\bibitem{7238334mlp}
Gurpreet Singh and Manoj Sachan.
\newblock Multi-layer perceptron (mlp) neural network technique for offline handwritten gurmukhi character recognition.
\newblock In {\em 2014 IEEE International Conference on Computational Intelligence and Computing Research}, pages 1--5, 2014.

\bibitem{soldaini2024dolmaopencorpustrillion}
Luca Soldaini, Rodney Kinney, Akshita Bhagia, Dustin Schwenk, David Atkinson, Russell Authur, Ben Bogin, Khyathi Chandu, Jennifer Dumas, Yanai Elazar, Valentin Hofmann, Ananya~Harsh Jha, Sachin Kumar, Li~Lucy, Xinxi Lyu, Nathan Lambert, Ian Magnusson, Jacob Morrison, Niklas Muennighoff, Aakanksha Naik, Crystal Nam, Matthew~E. Peters, Abhilasha Ravichander, Kyle Richardson, Zejiang Shen, Emma Strubell, Nishant Subramani, Oyvind Tafjord, Pete Walsh, Luke Zettlemoyer, Noah~A. Smith, Hannaneh Hajishirzi, Iz~Beltagy, Dirk Groeneveld, Jesse Dodge, and Kyle Lo.
\newblock Dolma: an open corpus of three trillion tokens for language model pretraining research, 2024.

\bibitem{Kocetkov2022TheStack}
Denis Kocetkov, Raymond Li, Loubna Ben~Allal, Jia Li, Chenghao Mou, Carlos Muñoz~Ferrandis, Yacine Jernite, Margaret Mitchell, Sean Hughes, Thomas Wolf, Dzmitry Bahdanau, Leandro von Werra, and Harm de~Vries.
\newblock The stack: 3 tb of permissively licensed source code.
\newblock {\em Preprint}, 2022.

\bibitem{penedo2024thefineweb}
Guilherme Penedo, Hynek Kydl{\'\i}{\v{c}}ek, Loubna~Ben allal, Anton Lozhkov, Margaret Mitchell, Colin Raffel, Leandro~Von Werra, and Thomas Wolf.
\newblock The fineweb datasets: Decanting the web for the finest text data at scale.
\newblock In {\em The Thirty-eight Conference on Neural Information Processing Systems Datasets and Benchmarks Track}, 2024.

\bibitem{schuhmann2022laion5bopenlargescaledataset}
Christoph Schuhmann, Romain Beaumont, Richard Vencu, Cade Gordon, Ross Wightman, Mehdi Cherti, Theo Coombes, Aarush Katta, Clayton Mullis, Mitchell Wortsman, Patrick Schramowski, Srivatsa Kundurthy, Katherine Crowson, Ludwig Schmidt, Robert Kaczmarczyk, and Jenia Jitsev.
\newblock Laion-5b: An open large-scale dataset for training next generation image-text models, 2022.

\bibitem{laioncoco}
LAION.
\newblock Laion-coco: 600m synthetic captions from laion-2b-en.
\newblock \url{https://laion.ai/blog/laion-coco/}.

\bibitem{chen2015microsoftcococaptionsdata}
Xinlei Chen, Hao Fang, Tsung-Yi Lin, Ramakrishna Vedantam, Saurabh Gupta, Piotr Dollar, and C.~Lawrence Zitnick.
\newblock Microsoft coco captions: Data collection and evaluation server, 2015.

\bibitem{sidorov2019textcaps}
Oleksii Sidorov, Ronghang Hu, Marcus Rohrbach, and Amanpreet Singh.
\newblock Textcaps: a dataset for image captioningwith reading comprehension.
\newblock 2020.

\bibitem{pixparse_pdfa_eng_wds}
PixParse Team.
\newblock pdfa-eng-wds dataset.
\newblock https://huggingface.co/datasets/pixparse/pdfa-eng-wds, 2024.
\newblock Accessed: 2025-02-22.

\bibitem{mathew2021docvqa}
Minesh Mathew, Dimosthenis Karatzas, and CV~Jawahar.
\newblock Docvqa: A dataset for vqa on document images.
\newblock In {\em Proceedings of the IEEE/CVF winter conference on applications of computer vision}, pages 2200--2209, 2021.

\bibitem{zhu2016visual7wgroundedquestionanswering}
Yuke Zhu, Oliver Groth, Michael Bernstein, and Li~Fei-Fei.
\newblock Visual7w: Grounded question answering in images, 2016.

\bibitem{8978122ocrvqa}
Anand Mishra, Shashank Shekhar, Ajeet~Kumar Singh, and Anirban Chakraborty.
\newblock Ocr-vqa: Visual question answering by reading text in images.
\newblock In {\em 2019 International Conference on Document Analysis and Recognition (ICDAR)}, pages 947--952, 2019.

\bibitem{xu2024llavacotletvisionlanguage}
Guowei Xu, Peng Jin, Hao Li, Yibing Song, Lichao Sun, and Li~Yuan.
\newblock Llava-cot: Let vision language models reason step-by-step, 2024.

\bibitem{gadre2023datacompsearchgenerationmultimodal}
Samir~Yitzhak Gadre, Gabriel Ilharco, Alex Fang, Jonathan Hayase, Georgios Smyrnis, Thao Nguyen, Ryan Marten, Mitchell Wortsman, Dhruba Ghosh, Jieyu Zhang, Eyal Orgad, Rahim Entezari, Giannis Daras, Sarah Pratt, Vivek Ramanujan, Yonatan Bitton, Kalyani Marathe, Stephen Mussmann, Richard Vencu, Mehdi Cherti, Ranjay Krishna, Pang~Wei Koh, Olga Saukh, Alexander Ratner, Shuran Song, Hannaneh Hajishirzi, Ali Farhadi, Romain Beaumont, Sewoong Oh, Alex Dimakis, Jenia Jitsev, Yair Carmon, Vaishaal Shankar, and Ludwig Schmidt.
\newblock Datacomp: In search of the next generation of multimodal datasets, 2023.

\bibitem{li2023mimicitmultimodalincontextinstruction}
Bo~Li, Yuanhan Zhang, Liangyu Chen, Jinghao Wang, Fanyi Pu, Jingkang Yang, Chunyuan Li, and Ziwei Liu.
\newblock Mimic-it: Multi-modal in-context instruction tuning, 2023.

\bibitem{yu2024rlaifv}
Tianyu Yu, Haoye Zhang, Yuan Yao, Yunkai Dang, Da~Chen, Xiaoman Lu, Ganqu Cui, Taiwen He, Zhiyuan Liu, Tat-Seng Chua, and Maosong Sun.
\newblock Rlaif-v: Aligning mllms through open-source ai feedback for super gpt-4v trustworthiness.
\newblock {\em arXiv preprint arXiv:2405.17220}, 2024.

\bibitem{lu2022learnscienceqa}
Pan Lu, Swaroop Mishra, Tony Xia, Liang Qiu, Kai-Wei Chang, Song-Chun Zhu, Oyvind Tafjord, Peter Clark, and Ashwin Kalyan.
\newblock Learn to explain: Multimodal reasoning via thought chains for science question answering.
\newblock In {\em The 36th Conference on Neural Information Processing Systems (NeurIPS)}, 2022.

\bibitem{bai2025qwen25vltechnicalreport}
Shuai Bai, Keqin Chen, Xuejing Liu, Jialin Wang, Wenbin Ge, Sibo Song, Kai Dang, Peng Wang, Shijie Wang, Jun Tang, Humen Zhong, Yuanzhi Zhu, Mingkun Yang, Zhaohai Li, Jianqiang Wan, Pengfei Wang, Wei Ding, Zheren Fu, Yiheng Xu, Jiabo Ye, Xi~Zhang, Tianbao Xie, Zesen Cheng, Hang Zhang, Zhibo Yang, Haiyang Xu, and Junyang Lin.
\newblock Qwen2.5-vl technical report, 2025.

\end{thebibliography}

\end{document}